\pdfoutput=1

\documentclass[11pt, twocolumn]{article}

\usepackage{ACL2023}
\usepackage{times}
\usepackage{amsmath}
\usepackage{amssymb}
\usepackage{amsfonts}
\usepackage{latexsym}
\usepackage{booktabs}
\usepackage{graphicx}
\usepackage{listings}
\usepackage{rotating} 
\usepackage{hyperref}
\usepackage[T1]{fontenc}

\usepackage[utf8]{inputenc}

\usepackage{microtype}

\usepackage{inconsolata}

\usepackage{xfrac} 

\usepackage{adjustbox} 
\usepackage{array} 

\newcolumntype{R}[2]{%
    >{\adjustbox{angle=#1,lap=(#2)-\width,valign=B}\bgroup}%
    c%
    <{\egroup}%
}
\newcommand*\rot{\multicolumn{1}{R{285}{1em}}}
\newcommand*\rotb{\multicolumn{1}{R{285}{1em}|}}

\usepackage{ifthen}

\newboolean{showcomments}
\setboolean{showcomments}{true} 

\usepackage{xcolor}

\definecolor{codegreen}{rgb}{0,0.6,0}
\definecolor{codegray}{rgb}{0.5,0.5,0.5}
\definecolor{codepurple}{rgb}{0.58,0,0.82}
\definecolor{backcolour}{rgb}{0.95,0.95,0.92}

\lstdefinestyle{mystyle}{
    backgroundcolor=\color{backcolour},   
    commentstyle=\color{codegreen},
    keywordstyle=\color{magenta},
    numberstyle=\tiny\color{codegray},
    stringstyle=\color{codepurple},
    basicstyle=\ttfamily\footnotesize,
    breakatwhitespace=false,         
    breaklines=true,                 
    captionpos=b,                    
    keepspaces=true,                 
    numbers=left,                    
    numbersep=5pt,                  
    showspaces=false,                
    showstringspaces=false,
    showtabs=false,                  
    tabsize=2
}

\title{Improving {\sc Instruct} Models for Free: A Study on Partial Adaptation}

\author{Ozan \.{I}rsoy$^1$, Pengxiang Cheng$^1$, Jennifer L. Chen$^2$\Thanks{ Work done while at Bloomberg.}, \\  \bf Daniel Preo\c{t}iuc-Pietro$^1$, Shiyue Zhang$^1$, Duccio Pappadopulo$^1$\Thanks{ Author ordering chosen at random.} \\
Bloomberg$^1$ \quad Independent Researcher$^2$ \\
  \texttt{\{oirsoy, pcheng134, dpreotiucpie, szhang1061, dpappadopulo\}@bloomberg.net}\\
  \texttt{jc4686@columbia.edu}
  }

\begin{document}
\maketitle
\begin{abstract}
Instruct models, obtained from various instruction tuning or post-training steps, are commonly deemed superior and more usable than their base counterpart. While the model gains instruction following ability, instruction tuning may lead to forgetting the knowledge from pre-training or it may encourage the model to become overly conversational or verbose. This, in turn, can lead to degradation of  in-context few-shot learning performance. In this work, we study the performance trajectory between base and instruct models by scaling down the strength of instruction-tuning via the partial adaption method. We show that, across several model families and model sizes, reducing the strength of instruction-tuning results in material improvement on a few-shot in-context learning benchmark covering a variety of classic natural language tasks. This comes at the cost of losing some degree of instruction following ability as measured by AlpacaEval. Our study shines light on
the potential trade-off between in-context learning and instruction following abilities that is worth considering in practice.

\end{abstract}

\section{Introduction}

\begin{figure*}[ht]
    \centering
    \includegraphics[width=\textwidth]{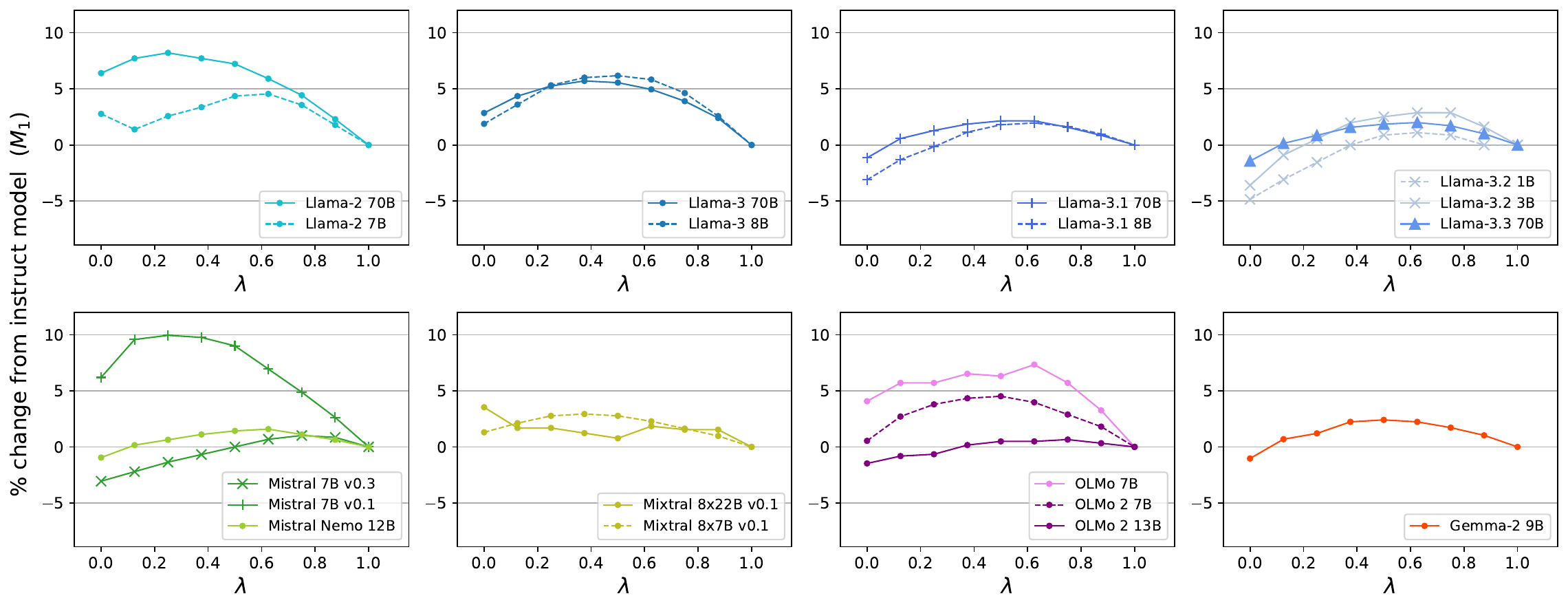}
    \caption{Performance on the in-context learning benchmark: fractional difference (percent value) between the performance of each partially adapted model $M_\lambda$ and the instruct baseline $M_1$ for all the models we have tested. }
    \label{fig:curves}
\end{figure*}

Training Large Language Models (LLMs) involves multiple steps, broadly categorized into pre-training and post-training. In pre-training, the \emph{base} model acquires the bulk of its knowledge through the next-token prediction objective. Post-training usually involves supervised fine-tuning (SFT) and multiple rounds of reinforcement learning from human feedback (RLHF), resulting in an \emph{instruct} model that is better at following instructions and more aligned with user goals. 

However, both SFT and RLHF, to some degree, encourage the model to produce long and conversational responses. This may be an unwanted feature when testing on extractive and/or structured natural language processing (NLP) tasks such as classification, named entity recognition, or extractive question answering. In these cases, the responses need to be concise and exact, and any additional chattiness creates issues in parsing the responses. Before instruct models became available, this need was fulfilled reasonably well by the emergent few-shot in-context learning (ICL) abilities of the base model~\cite{wei2022emergent}. Few previous studies touch on the pros and cons of base and instruct models. One example is \cite{cuconasu2024tale} which shows how base models work better than instruct models on RAG-related tasks.

Our work aims to fill this gap and thoroughly explores the performance trajectory between base and instruct models. In order to study the learning dynamics between base and instruct models, we need access to the model checkpoints saved during instruct tuning, which are rarely available, especially for best performing open-weight models. Therefore as a surrogate of this~\cite{na-etal-2024-scalable}, we resort to a simple training-free technique, \emph{partial adaptation (or PAd)}~\cite{fleshman2024re},  to scale the instruction-tuning strength in a post-hoc manner. Specifically, we create in-between models by partially adapting the base model (with weights $\mathbf{W_B}$) to instruct (with weights $\mathbf W_I$): $M_\lambda$ with weights $\mathbf{W_B} + \lambda \mathbf{A} \;(\lambda \in [0,1])$ where $\mathbf A\equiv \mathbf W_I-\mathbf W_B$. Hence, $M_0$ is the base model and $M_1$ is the instruct model  (see Section~\ref{sec:methods} for more details).

Using 18 open-weight LLMs, we evaluate these partially adapted models on a benchmark containing 21 classic NLP tasks using few-shot in-context learning. We find that, for all models, the best performance is always achieved when $\lambda < 1$, i.e., when instruction tuning strength is scaled down. And the optimal choice of $\lambda$ leads to a few percent points improvement with respect to both the base and instruct models. 

However, perhaps not surprisingly, we also find that once evaluated on an instruction following benchmark, AlpacaEval 2.0~\cite{alpacaeval}, the best partially adapted models selected by the ICL benchmark consistently under-perform their fully instruction tuned counterparts. Nonetheless, especially for models of larger sizes, we can oftentimes find a $\lambda < 1$, for which the AlpacaEval performance shows little to no drop, yet there is still a gain in the ICL benchmark.

In summary, through this comprehensive analysis, we demonstrate that the best ICL model is not necessarily the instruct model. We believe partial adaptation represents a training-free yet effective option worth exploring when dealing with ICL tasks that are structured, more extractive in nature, or requiring shorter answers. We hope our study highlights the opportunities and can inspire future work in better understanding the learning dynamics in LLM post-training.

\section{Preliminary: Partial Adaptation}
\label{sec:methods}

\citet{fleshman2024re} propose that the contribution of LLM post-training can be isolated by simply differencing the weights of the instruct and base model, $ \mathbf A\equiv \mathbf W_I-\mathbf W_B$. $\mathbf A$ can be seen as an adapter to be applied on top of the base model and the strength of the adapter can be adjusted in the form of $\mathbf{W_B} + \lambda \mathbf{A} \;(\lambda \in [0,1])$. This technique is called \emph{partial adaptation} (PAd), with the implied meaning as \emph{partially adapting the base model to instruction following}. In fact, in one single experiment, \citet{fleshman2024re} also showed that partial adaptation leads to improvement on a zero-shot QA task to support their conjecture that instruction-tuning likely degrades knowledge from pretraining. We are inspired by this observation and conduct thorough analysis across models and datasets in this paper. 

The partially adapted model can also be viewed as the weighted average between base and instruct models. Hence, we consider a new model $M_\lambda$ with weights $(1-\lambda) \mathbf{W}_B + \lambda  \mathbf{W}_I$, so that $M_0$ and $ M_1$ correspond to the base and instruct models respectively. Open-weight models that we consider are listed in Table ~\ref{tab:results}.\footnote{For all of the models, except Mixtral 8$\times$22B, the embedding lookup tables of the base and instruct versions are aligned, so merging is straightforward. For Mixtral 8$\times$22B, there are additional special tokens in the vocabulary of the instruct model.
We take care of this by applying $\lambda=1$ for those weights that are only present in the instruct model.} In practice, we enumerate $\lambda$ from $\{0, \frac{1}{8}, \frac{2}{8}, \frac{3}{8}, \frac{4}{8}, \frac{5}{8}, \frac{6}{8}, \frac{7}{8}, 1\}$.

\section{Evaluation Benchmarks}
\label{sec:data}

We evaluate partially adapted models on two benchmarks for testing ICL and instruction following performance respectively.

\subsection{In-Context Learning Benchmark}

Our primary goal is to measure performance on few-shot in-context learning. We assemble a benchmark of various classic NLP tasks to test a variety of natural language abilities. The composition of the benchmark is shown in Table~\ref{tab:benchmark} and described in detail in Appendix~\ref{sec:app:datasets}. 
We particularly include tasks from the financial domain because classic structured NLP tasks (classification, named entity recognition, extractive QA) are common in financial data analysis. Each dataset is tested in a few-shot manner, where the number of shots is displayed in Table~\ref{tab:benchmark}. Shot selection is random and done independently for each example.

Depending on the dataset, evaluation proceeds in one of three possible ways (more details in Appendix~\ref{sec:app:tasks}). For multiple choice (MC) datasets, we use the model to score each of the possible answers using likelihood and pick the highest ranking one. As a variation of this, fast multiple choice (FMC), instead of scoring each response, the model is prompted with them as a bulleted list (in MMLU format~\cite{hendrycks2021measuringmassivemultitasklanguage}) and only the individual tokens corresponding to the bullets ($A$, $B$, $C$, ...) are scored and ranked. Finally for generation (G) datasets, the model generates a completion which is then parsed and compared to the ground truth answer.\footnote{Note that both MC and FMC are standard evaluation protocols for multiple choice tasks used by \href{https://github.com/mosaicml/llm-foundry/blob/main/llmfoundry/eval/metrics/nlp.py\#L296}{LLM-foundry} and \href{https://github.com/hendrycks/test/blob/master/evaluate.py\#L89}{MMLU}.}

When a single dataset is evaluated in multiple ways (different prompts or different evaluation styles: MC vs. FMC vs. G), we aggregate these individual scores by taking their maximum. All metric sores are in a scale of 0 to 100. Therefore, we are able to \emph{average} dataset-level scores into one single model-level score. 
More details about the templates and metrics that we use in our evaluation protocol are presented in Appendix~\ref{sec:app:templates} and~\ref{sec:app:metrics}.

\subsection{AlpacaEval}
Instruction following is a broad concept. In this work, we refer to it as the model's ability to answer open-ended questions from users, as exemplified by Chatbot Arena~\cite{chatbot_arena}.
Here, we test on AlpacaEval 2.0~\cite{alpacaeval}, which has a Spearman correlation of 0.98 with Chatbot Arena while being cost-efficient. 
For each value of $\lambda$, we obtain the length-controlled win-rate of $M_\lambda$ against GPT-4 Preview (11/06)~\cite{alpacaeval_git} judged by GPT-4o.\footnote{The GPT-4o version that we use is the May 2024 one.} 

\begin{table}[t!]
\small
\begin{center}

\begin{tabular}{ llllll } 
 \toprule
Model  & Base/Inst. & Best$^{\lambda^*}$ & $\delta_{\textrm{wr}}$  \\
\midrule
Llama-2 7B & 51.9/50.5 & 52.8$^{5/8}$ & $-$4.35\\ 
Llama-2 70B & 64.8/60.9 & 65.9$^{2/8}$ & $-$16.64\\ 
Llama-3 8B & 59.4/58.3 & 61.9$^{4/8}$ & $-$15.81\\ 
Llama-3 70B & 68.5/66.6 & 70.4$^{3/8}$ & $-$6.02\\ 
Llama-3.1 8B & 59.3/61.2 & 62.4$^{5/8}$ & $-$5.58\\ 
Llama-3.1 70B & 69.0/69.8 & 71.3$^{4/8}$ & $-$5.30\\ 
Llama-3.2 1B & 43.2/45.4 & 45.9$^{5/8}$ & $-$8.93\\
Llama-3.2 3B & 53.6/55.6 & 57.2$^{5/8}$ & $-$8.89\\ 
Llama-3.3 70B & 69.0/70.0 & 71.4$^{5/8}$ & $-$0.93\\ 
\midrule
Mistral 7B v0.1 & 56.6/53.3 & 58.6$^{2/8}$ & $-$6.73\\
Mistral 7B v0.3 & 57.1/58.9 & 59.5$^{6/8}$ & $-$1.57\\ 
Mistral Nemo 12B & 62.5/63.1 & 64.1$^{5/8}$ & $-$5.70\\ 
\midrule
Mixtral 8x7B v0.1 & 62.2/61.4 & 63.2$^{3/8}$ & $-$14.48\\  
Mixtral 8x22B v0.1 & 67.4/65.1 & 67.4$^{0/8}$ & ~~~~NA \\
\midrule
Gemma-2 9B & 57.6/58.2 & 59.6$^{4/8}$ & $-$6.52\\ 
\midrule
OLMo 7B 0724 & 51.1/49.1 & 52.7$^{5/8}$ & $-$6.79\\ 
OLMo 2 7B 1124 & 55.7/55.4 & 57.9$^{4/8}$ & $-$7.41\\ %
OLMo 2 13B 1124 & 60.2/61.1 & 61.5$^{6/8}$ & $-$3.98\\ %
\bottomrule
\end{tabular}

\end{center}
\caption{
For each of the models 
(LLama-2~\cite{touvron2023llama2openfoundation}, Llama-3~\cite{llama3, llama31, llama32, llama33}, Mistral~\cite{jiang2023mistral}, Mistral-NeMo~\cite{mistralnemo}, Mixtral~\cite{jiang2024mixtral, mixtral822}, Gemma-2~\cite{team2024gemma}, OLMo~\cite{olmo}, and OLMo-2~\cite{olmo2}) 
in the first column we report the base and instruct baseline performance on the benchmark, together with the best performance obtained by varying $\lambda$ and the best value $\lambda^*$ at which peak performance is achieved. The last column reports the \emph{absolute} change in win rate for the best PAd model with respect to the instruct version as determined by AlpacaEval 2.0. NA is because $\lambda^*=0$ and we don't evaluate AlpacaEval on the base model when chat template does not exist.
}
\label{tab:results}

\end{table}

\begin{figure*}[h!]
    \centering
    \includegraphics[width=\textwidth]{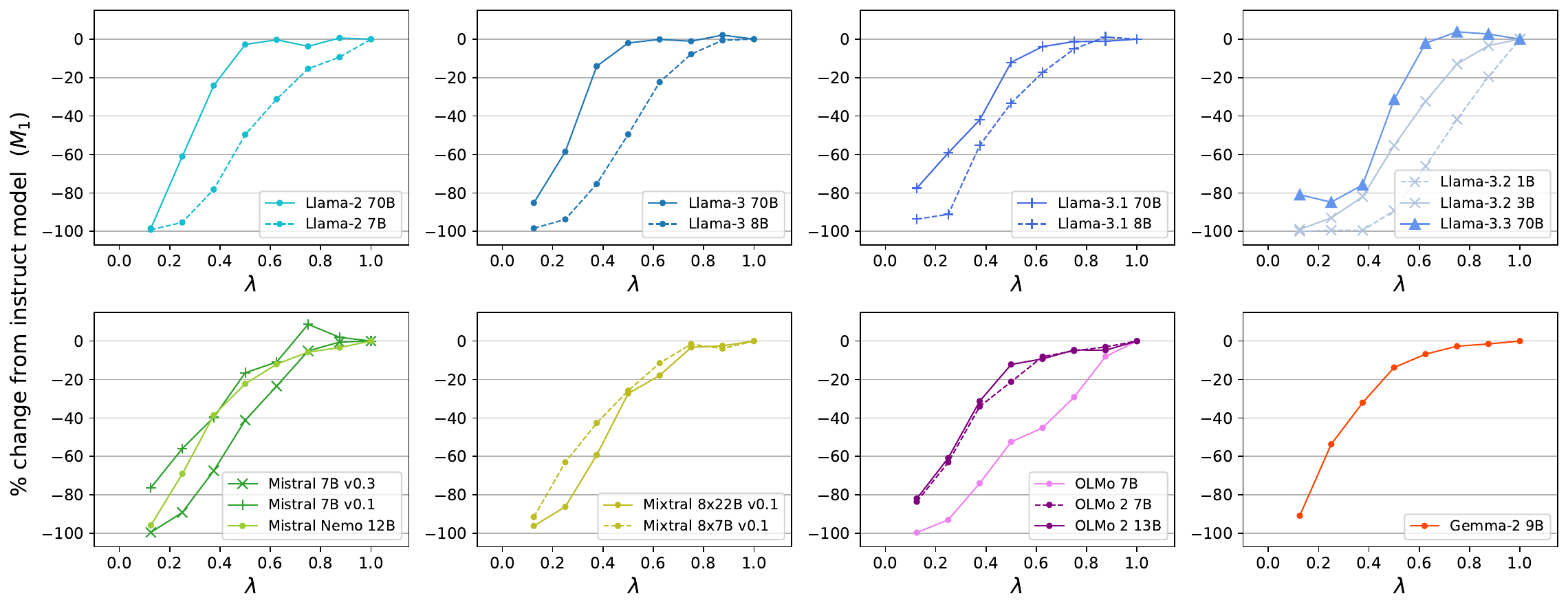}
    \caption{Performance on AlpacaEval 2.0: fractional difference (percent value) between the length-controlled win rate of each partially adapted model $M_\lambda$ and the instruct baseline $M_1$ against GPT-4 Preview (11/06).}
    \label{fig:curves_alpaca}
\end{figure*}

\section{Results}
Figure~\ref{fig:curves} and Figure~\ref{fig:curves_alpaca} illustrate the relative performance change of each partially adapted model $M_\lambda$ against the instruct model $M_1$ on ICL and AlpacaEval benchmark, respectively. And Figure~\ref{fig:curves_abs} and Figure~\ref{fig:curves_alpaca_abs} in Appendix~\ref{sec:app:results} shows the corresponding absolute values. We summarize the absolute performance of base/instruct models and the best partially adapted models as well as the best $\lambda^*$ in Table~\ref{tab:results}. 

\textbf{The best ICL performance is always achieved by \emph{less} instruction-tuned models.} As shown by Figure~\ref{fig:curves}, for all 18 models, the peak of the curves is reached when $\lambda < 1$. It means scaling down instruction tuning strength to some degree enhances in-context learning ability. In addition, for 17 out of 18 models, except for Mixtral 8x22B, PAd improves ICL performance over both base and instruct models. For 15 out of 18 models, this improvement is greater than $0.5$. The largest improvement we observe is $2.5$ on Llama-3 8B. The best $\lambda$ is oftentimes between 0.5 to 0.6. 
Similar trends are evident at the individual dataset level (Table~\ref{tab:results_nonagg}).

\textbf{The improvement on ICL is at the cost of losing some instruction following abilities} 
as measured by the AlpacaEval 2.0 win rate shown in Figure~\ref{fig:curves_alpaca} and the last column of Table~\ref{tab:results}. In Table~\ref{tab:results}, 
$\delta_{\textrm{wr}}\equiv {\textrm{wr}}_{M_{\lambda^*}} - {\textrm{wr}}_{M_1}$
represents the \emph{absolute} difference in win rate between the best PAd model for ICL ($M_{\lambda^*}$) and the instruct version ($M_1$). As shown in Figure~\ref{fig:curves_alpaca}, the best win rate is mostly achieved by the instruct model, except for a few cases where a marginally higher win rate is achieved when  $0.6 < \lambda < 1$.

\textbf{ICL can be improved with a small drop of instruction following abilities.}
We notice that for many models, especially the larger ones, the win-rate curve saturates to the instruct value for $\lambda$ values well below 1. This implies that there are values of $\lambda$, in the range $\lambda^*\leq \lambda < 1$, where the AlpacaEval 2.0 performance does not drop significantly, yet there is still a gain on the ICL benchmark due to PAd. For instance, by allowing a relative win-rate decrease of at most 1\% from the instruct model on AlpacaEval 2.0, we can get a $+5.9\%$ relative improvement on the ICL benchmark performance for Llama-2 70B ($\lambda=0.625$), $+4.9\%$ for Llama-3 70B ($\lambda=0.625$), $+1.7\%$ for Llama-3.3 70B ($\lambda=0.75$).

\section{Conclusion and Future Work}
In this work, we study the performance trajectory between base and instruct models for 18 LLMs via the training-free partial adaptation method~\cite{fleshman2024re}. We find that scaling down instruction tuning strength can benefit in-context learning tasks for all models across 21 datasets. However, this improvement is at the cost of losing instruction following ability. 

Nonetheless, the observation that instruction following performance for larger models is not very sensitive to $\lambda$ when $\lambda \lesssim 1$ suggests that slightly scaling down instruction tuning strength would consistently be beneficial. Hence, it would make sense to apply PAd at the end of post-training (e.g., replacing $M_1$ with $M_{\lambda^*}$) to further boost model performance. This might have already happened as Llama 3.3~\cite{llama33} used an annealing technique to average model checkpoints, and we also observed that PAd boosts Llama 2 ICL performance much more than Llama 3.3.  

Future work can focus on better understanding why PAd improves ICL performance by studying its impact on each stage of supervised fine-tuning or RL. Another avenue of investigation is a thorough comparison of the training dynamics during instruction tuning with the model trajectory 
 defined by varying $\lambda$ in PAd. It has been suggested that the latter may indeed replicate the full training dynamics~\cite{na-etal-2024-scalable}.

\section*{Limitations}

Our in-context learning benchmark is a collection of 21 common datasets  spanning 6 broad types of tasks. The collection may however not be fully representative of the model ICL performance or its performance on other specific tasks. Similarly, we benchmark instruction following ability on AlpacaEval 2.0~\cite{alpacaeval}, which has a Spearman correlation of 0.98 with Chatbot Arena. However, it may not be fully representative of the model true instruction following performance. Further, we limit our study to models primarily trained on English data and tasks in English, hence we leave testing the generalizability to other languages and multilingual models to future work.

\section*{Acknowledgements}
We thank Steven Lu and Shijie Wu for their contributions to developing the in-context learning benchmark.

\bibliography{custom}
\bibliographystyle{acl_natbib}

\onecolumn
\appendix
\section{In-context Learning Benchmark Details}
\subsection{Datasets}
\label{sec:app:datasets}

\begin{table*}[t!]
\footnotesize
\begin{center}
\resizebox{0.98\textwidth}{!}{
\begin{tabular}{ llllll } 
 \toprule
Capability & Domain & Dataset & Shots & Style & Size\\
\midrule
World Knowledge  & General & MMLU \cite{hendrycks2021measuringmassivemultitasklanguage} & 5 & MC, FMC & 14042 \\
&& Trivia QA \cite{joshi-etal-2017-triviaqa} & 1 & G & 1105\\
&& Natural Questions \cite{47761} & 1 & G & 1032\\

\midrule
Commonsense Reasoning & General & PIQA \cite{bisk2020piqa} & 1 & MC & 1838\\
&& Winogrande \cite{sakaguchi2021winogrande} & 1  & MC & 1267\\
&& ARC Challenge \cite{clark2018think} & 1 & MC & 1172\\
&& HellaSwag \cite{zellers-etal-2019-hellaswag} & 1 & MC & 10042\\
 \midrule
Language Processing and Understanding
& General & BBH (NLP) \cite{suzgun-etal-2023-challenging} & 3 & G, MC & 3000\\
\cmidrule{2-6}
& Finance & FiQA (SA) \cite{shah2022flue} & 5 & MC, FMC & 235\\
&& FPB (SA) \cite{shah2022flue} & 5 & MC, FMC & 970\\
&& Headline \cite{shah2022flue} & 5 & MC, FMC & 20547\\
&& Flue (NER) \cite{shah2022flue} & 20 & G & 98\\
 \midrule
Symbolic and Logical Problem Solving
& General & BBH (Algo) \cite{suzgun-etal-2023-challenging} & 3 & G, MC 
& 3000\\
&& DROP \cite{dua2019drop} & 1 & G & 1000 \\
\cmidrule{2-6}
&Finance& TAT-QA \cite{zhu-etal-2021-tat} & 1 & G & 1668\\
&& Pacific \cite{deng-etal-2022-pacific} & 1 & G & 1982\\
 \midrule
Reading Comprehension
& General & SQuAD \cite{rajpurkar-etal-2016-squad} & 2 & G & 1000\\

&& QuAC \cite{choi-etal-2018-quac} & 2 & G & 1000\\
\cmidrule{2-6}
&Finance & ConvFinQA \cite{chen-etal-2022-convfinqa} & 1 & G & 5932\\
 \midrule
Retrieval-augmented Generation (RAG)
& General & Natural Questions + Wiki   \cite{47761} & 1 & G & 1105\\
&& Trivia QA + Wiki  \cite{joshi-etal-2017-triviaqa} & 1& G & 1032\\

\bottomrule
\end{tabular}
}
\end{center}
\caption{
A complete list of the datasets composing our in-context few-shot 
learning evaluation benchmark. The last column (Size) shows the number of examples in each dataset. }
\label{tab:benchmark}
\end{table*}

Table~\ref{tab:benchmark} lists the datasets we used to build the ICL benchmark, which are organized in a taxonomy according to the ability they are supposed to test and the domain they are operating on. 
\begin{itemize}
    \item \textbf{World knowledge}: we include the widely used Massive Multitask Language Understanding (MMLU) benchmark~\cite{hendrycks2021measuringmassivemultitasklanguage} and two open-domain QA tasks, Trivia QA~\cite{joshi-etal-2017-triviaqa} and Natural Questions~\cite{47761}.
    \item \textbf{Commonsense reasoning}: four datasets (PIQA~\cite{bisk2020piqa}, Winogrande~\cite{sakaguchi2021winogrande}, ARC Challenge~\cite{clark2018think}, and HellaSwag~\cite{zellers-etal-2019-hellaswag}) to test different types of commonsense reasoning ability of the model. 
    \item \textbf{Language processing and understanding}: we include five classic language processing or understanding tasks. BBH (NLP) are NLP tasks from Big Bench Hard~\cite{suzgun-etal-2023-challenging}, e.g., movie recommendation. FiQA (SA) and FFB (SA) are two sentiment analysis tasks, Headline is a headline classification task, Flue (NER) is a named entity recognition task, and all these four datasets are from FLUE (Financial Language Understanding Evaluation) benchmark~\cite{shah2022flue}.   
    \item \textbf{Symbolic and logical problem solving}: BBH (Algo) contains algorithmic tasks (e.g., Boolean expressions) from Big Bench Hard~\cite{suzgun-etal-2023-challenging}. DROP~\cite{dua2019drop} is a discrete reasoning QA dataset. TAT-QA~\cite{zhu-etal-2021-tat} and Pacific~\cite{deng-etal-2022-pacific} are two financial table QA tasks. 
    \item \textbf{Reading comprehension}: SQuAD~\cite{rajpurkar-etal-2016-squad} and QuAC~\cite{choi-etal-2018-quac} are two general-domain reading comprehension QA datasets, and ConvFinQA~\cite{chen-etal-2022-convfinqa} is a financial QA task. 
    \item \textbf{Retrieval-augmented generation (RAG)}: we use questions from Natural Questions~\cite{47761} and Trivia QA~\cite{joshi-etal-2017-triviaqa} to retrieve passages from Wikipedia, which creates two RAG evaluation tasks.
\end{itemize}

\subsection{Evaluation Tasks}
\label{sec:app:tasks}

\begin{table}[t!]
\small
\begin{center}
\begin{tabular}{ llll } 
 \toprule
Template & Dataset & Style & Metric \\
\midrule
\texttt{mmlu\_joint.j2} & MMLU & FMC & Accuracy\\
\texttt{mmlu\_separate.j2} & MMLU & MC& Accuracy\\
\midrule
\texttt{instruct\_qa.j2} & BBH & G& Accuracy\\
\texttt{bbh\_separate.j2} & BBH & MC& Accuracy\\
\midrule
\texttt{sa\_t4.j2} & FBP (SA) & MC& Weighted F1\\
\texttt{sa\_t4\_opt.j2} & FBP (SA) & MC& Weighted F1\\
\texttt{sa\_t4\_joint.j2} & FBP (SA) & FMC& Weighted F1\\
\midrule
\texttt{ner\_inline.j2} & Flue (NER) & G& F1\\
\midrule
\texttt{simple\_qa.j2} & QuAC & G& String F1\\
& TAT-QA & G& Fin QA F1\\
& DROP & G& String F1\\
& ConvFinQA & G& Fin QA Accuracy\\
& SQuAD & G& String F1\\
& Natural Questions & G& String F1\\
& Trivia QA & G& String F1\\
\midrule
\texttt{simple\_qa\_new.j2} & Natural Questions + Wiki & G & String F1\\
& Trivia QA + Wiki & G & String F1\\
\midrule
\texttt{simple\_qa\_mc.j2} & ARC Challenge & MC& Accuracy\\
\cmidrule{2-4}
\texttt{simple\_qa\_mc\_opt.j2} & Headline & MC& Average Weighted F1\\
\texttt{simple\_qa\_mc\_joint.j2} & Headline & FMC& Average Weighted F1\\
\midrule
\texttt{asa\_t4.j2} & FiQA & MC& Weighted F1 \\
\texttt{asa\_t4\_opt.j2} & FiQA & MC& Weighted F1\\
\texttt{asa\_t4\_joint.j2} & FiQA & FMC& Weighted F1\\
\midrule
\texttt{pacific.j2} & Pacific & G& Fin QA F1\\
\midrule
\texttt{mc\_concat.j2} & HellaSwag & MC& Accuracy\\
& Winogrande & MC& Accuracy\\
& PIQA & MC& Accuracy\\
\bottomrule
\end{tabular}
\end{center}
\caption{
Templates used for to evaluate each of the datasets. We also show the metrics used to evaluate the different datasets. If a single dataset is evaluated multiple times using different templates or styles, the final scores are aggregated by taking their maximum.}
\label{tab:template_metrics}

\end{table}

The in-context benchmark is composed of three categories of tasks.
\begin{itemize}
    \item Multiple choice (MC): For multiple choice datasets, we use the model to score the likelihood of each of the possible choices $c\in \mathcal C$ and pick the highest ranking one, $c^*$,
    \begin{equation}
       c^* \equiv \underset{c\in \mathcal C}{\textrm{argmax}} ~ \mathbb{P}(c\,|\,{\textrm{prompt}}) / N(c)
    \end{equation}
    $N(c)$ is a possibly choice dependent normalization that we use to ameliorate possible biases of the model likelihood~\cite{mcbias}. We consider 3 possibilities for $N$
    \begin{align}
    N_{\textrm{base}}(c)&= 1\\
    N_{\textrm{length}}(c)&= |{\texttt{tokens}(c)}|\\
    N_{\textrm{prior}}(c)& = \mathbb{P}(\textrm{prefix})\label{prefix_cal}
    \end{align}
    where ${\texttt{tokens}(c)}$ is the list of tokens representing $c$ and $\mathbb{P}(\textrm{prefix})$ is the probability that the model assigns to a generic prefix that does not depend on $c$, for instance the string "Answer: " (see Appendix~\ref{sec:app:templates} for details). We calculate accuracy or F1 score for each of these choices of $N$ and we aggregate the final results by taking the maximum across these scores.
    \item Fast multiple choice (FMC): Similar to MC, but instead of asking the model to score each possible response, the model is shown the possible choices as a bulleted list (in MMLU format~\cite{hendrycks2021measuringmassivemultitasklanguage}) and only the individual tokens corresponding to the bullets ($A$, $B$, $C$, ...) are scored and ranked
    \begin{equation}
       c^*\equiv \underset{c\in \{A, \,B,\, C,\, ...\}}{\textrm{argmax}} ~\mathbb{P}(c\,|\,{\textrm{prompt}})    
    \end{equation}
    \item Generation (G): The model generates a completion which is then parsed and compared to the ground truth answer. Evaluation metrics include string-F1 and Exact Match. The full list of evaluation metrics is shown in Table~\ref{tab:template_metrics} and described in Appendix~\ref{sec:app:metrics}.
\end{itemize}

\subsection{Templates}
\label{sec:app:templates}

In this section we report the templates that we use in our experiments. All of them are displayed in \texttt{jinja2} format.

In some of the templates below (\texttt{mmlu\_separate.j2}, \texttt{bbh\_separate.j2}, \texttt{sa\_t4.j2}, \texttt{sa\_t4\_opt.j2}, \texttt{simple\_qa\_mc.j2}, \texttt{simple\_qa\_mc\_opt.j2}, \texttt{asa\_t4.j2}, \texttt{asa\_t4\_opt.j2}) the separator string \texttt{|||} appears. This is used to perform calibration following Eq.~\ref{prefix_cal}: the full template is obtained by replacing \texttt{|||} with the empty string, and the prefix appearing in Eq.~\ref{prefix_cal} is obtained by splitting the prompt at \texttt{|||}:
\begin{lstlisting}
_, prefix = template.split("|||")
template = template.replace("|||", "")
\end{lstlisting}

\subsubsection*{\texttt{mmlu\_joint.j2}}
\begin{lstlisting}
{% set ENUM = 'ABCDEFGHIJKLM' %}The following are multiple choice questions (with answers) about {{subject}}.
{% for example in examples %}
Question: {{ example.question }}
{% for choice in example.choices %}({{ ENUM[loop.index0] }}) {{ choice }}
{% endfor %}Answer: ({{ ENUM[example.gold] }}) {{ example.choices[example.gold] }}
{% endfor %}
Question: {{ question }}
{% for choice in choices %}({{ ENUM[loop.index0] }}) {{ choice }}
{% endfor %}Answer:
\end{lstlisting}

\subsubsection*{\texttt{mmlu\_separate.j2}}
\begin{lstlisting}
The following are multiple choice questions (with answers) about {{subject}}.
{% for example in examples %}
Question: {{ example.question }}
Answer: {{ example.choices[example.gold] }}
{% endfor %}
Question: {{ question }}
|||Answer:
\end{lstlisting}

\subsubsection*{\texttt{instruct\_qa.j2}}
\begin{lstlisting}
{% for example in examples %}{{ example.question }}
Answer: {{ example.gold }}

{% endfor %}{{ question }}
Answer:
\end{lstlisting}

\subsubsection*{\texttt{bbh\_separate.j2}}
\begin{lstlisting}
{{instruction}}
{% for example in examples %}
Question: {{ example.question }}
Answer: {{ example.choices[example.gold] }}
{% endfor %}
Question: {{ question }}
|||Answer:
\end{lstlisting}

\subsubsection*{\texttt{sa\_t4.j2}}
\begin{lstlisting}
{% for ex in examples %}{{ ex.sentence }}
Question: what is the sentiment?
Answer: {{ ex.choices[ex.gold] }}

{% endfor %}{{ sentence }}
Question: what is the sentiment?
|||Answer:
\end{lstlisting}

\subsubsection*{\texttt{sa\_t4\_opt.j2}}
\begin{lstlisting}
{% for ex in examples %}{{ ex.sentence }}
Question: what is the sentiment?
Options:
{% for choice in ex.choices %}- {{ choice }}
{% endfor %}Answer: {{ ex.choices[ex.gold] }}

{% endfor %}{{ sentence }}
Question: what is the sentiment?
Options:
{% for choice in choices %}- {{ choice }}
{% endfor %}|||Answer:
\end{lstlisting}

\subsubsection*{\texttt{sa\_t4\_joint.j2}}
\begin{lstlisting}
{% set ENUM = 'ABCDEFGHIJKLM' %}{% for ex in examples %}{{ ex.sentence }}
Question: what is the sentiment?
{% for choice in ex.choices %}({{ ENUM[loop.index0] }}) {{ choice }}
{% endfor %}Answer: ({{ ENUM[ex.gold] }}) {{ ex.choices[ex.gold] }}

{% endfor %}{{ sentence }}
Question: what is the sentiment?
{% for choice in choices %}({{ ENUM[loop.index0] }}) {{ choice }}
{% endfor %}Answer:
\end{lstlisting}

\subsubsection*{\texttt{simple\_qa.j2}}
\begin{lstlisting}
{% for example in examples %}{{ example.question }}
Answer: {{ example.gold }}

{% endfor %}{{ question }}
Answer:
\end{lstlisting}

\subsubsection*{\texttt{simple\_qa\_new.j2}}
\begin{lstlisting}
{% for example in examples %}{{ example.sources|join('\n\n') }}

{{ example.question }}
Answer: {{ example.gold }}

{% endfor %}{{ sources|join('\n\n') }}

{{ question }}
Answer:
\end{lstlisting}

\subsubsection*{\texttt{simple\_qa\_mc.j2}}
\begin{lstlisting}
{% for example in examples %}{{ example.question }}
Answer: {{ example.choices[example.gold] }}

{% endfor %}{{ question }}
|||Answer:
\end{lstlisting}

\subsubsection*{\texttt{simple\_qa\_mc\_opt.j2}}
\begin{lstlisting}
{% for ex in examples %}{{ ex.question }}
Options:
{% for choice in ex.choices %}- {{ choice }}
{% endfor %}Answer: {{ ex.choices[ex.gold] }}

{% endfor %}{{ question }}
Options:
{% for choice in choices %}- {{ choice }}
{% endfor %}|||Answer:
\end{lstlisting}

\subsubsection*{\texttt{simple\_qa\_mc\_joint.j2}}
\begin{lstlisting}
{% set ENUM = 'ABCDEFGHIJKLM' %}{% for ex in examples %}{{ ex.question }}
{% for choice in ex.choices %}({{ ENUM[loop.index0] }}) {{ choice }}
{% endfor %}Answer: ({{ ENUM[ex.gold] }}) {{ ex.choices[ex.gold] }}

{% endfor %}{{ question }}
{% for choice in choices %}({{ ENUM[loop.index0] }}) {{ choice }}
{% endfor %}Answer:
\end{lstlisting}

\subsubsection*{\texttt{asa\_t4.j2}}
\begin{lstlisting}
{% for ex in examples %}{{ ex.sentence }}
Question: what is the sentiment on {{ ex.target }}?
Answer: {{ ex.choices[ex.gold] }}

{% endfor %}{{ sentence }}
Question: what is the sentiment on {{ target }}?
|||Answer:
\end{lstlisting}

\subsubsection*{\texttt{asa\_t4\_opt.j2}}
\begin{lstlisting}
{% for ex in examples %}{{ ex.sentence }}
Question: what is the sentiment on {{ ex.target }}?
Options:
{% for choice in ex.choices %}- {{ choice }}
{% endfor %}Answer: {{ ex.choices[ex.gold] }}

{% endfor %}{{ sentence }}
Question: what is the sentiment on {{ target }}?
Options:
{% for choice in choices %}- {{ choice }}
{% endfor %}|||Answer:
\end{lstlisting}

\subsubsection*{\texttt{asa\_t4\_joint.j2}}
\begin{lstlisting}
{% set ENUM = 'ABCDEFGHIJKLM' %}{% for ex in examples %}{{ ex.sentence }}
Question: what is the sentiment on {{ ex.target }}?
{% for choice in ex.choices %}({{ ENUM[loop.index0] }}) {{ choice }}
{% endfor %}Answer: ({{ ENUM[ex.gold] }}) {{ ex.choices[ex.gold] }}

{% endfor %}{{ sentence }}
Question: what is the sentiment on {{ target }}?
{% for choice in choices %}({{ ENUM[loop.index0] }}) {{ choice }}
{% endfor %}Answer:
\end{lstlisting}

\subsubsection*{\texttt{pacific.j2}}
\begin{lstlisting}
{% for example in examples %}{{ example.question }}
{{ example.gold }}

{% endfor %}{{ question }}
\end{lstlisting}

\subsubsection*{\texttt{mc\_concat.j2}}
\begin{lstlisting}
{% for example in examples %}{{ example.question }}{{ example.choices[example.gold] }}

{% endfor %}{{ question }}
\end{lstlisting}

\newpage
\subsection{Metrics}
\label{sec:app:metrics}
Table~\ref{tab:template_metrics} lists the metrics used to evaluate each dataset in our benchmark. 
\begin{itemize}
    \item \textbf{Accuracy}: For classification tasks, it checks whether the predicted label matches the gold label. For generation tasks, it checks whether the generated answer matches the gold answer. 
    \item \textbf{Weighted F1}: Calculate F1 scores for each class, and find their average weighted by support (the number of true instances for each class). 
    \item \textbf{F1}: This metric is only used for the Flue (NER) task. For each entity type, there are a list of gold entities and a list of model-generated entities. True positive is the number of overlaps between ground-truth and model generations. False positive is the number of entities that the model generates but are not ground-truth. False negative is the number of entities that are gold but the model does not generate.
    \item \textbf{String F1}: We use the same evaluation script from SQuAD~\cite{rajpurkar2016squad}, in which gold and generated answers are treated as two bags of words. String F1 is the F1 score computed between these two bags of words.
    \item \textbf{Fin QA F1}: This metric is the same as String F1, except for two cases. When the gold answer is a \emph{number}, we extract and convert the model generation to a number and check if it matches the gold number. When the gold answer is yes or no, we check if the first word of model generation matches the gold answer. 
    \item  \textbf{Fin QA Accuracy}: This metric is similar to Fin QA F1, except that we replace String F1 with String EM (Exact Match) because the answers are mostly short. 
    \item \textbf{Average Weighted F1}: This metric is used when there are multiple groups of multi-choice classification tasks. We compute the weighted F1 within each group and then take the average across groups.
\end{itemize}
All metric scores are in a scale of 0 to 100. Therefore, we are able to average dataset-level scores into one single model-level score. 

\newpage
\section{Additional Results}
\label{sec:app:results}

\begin{figure*}[h!]
    \centering
    \includegraphics[width=\textwidth]{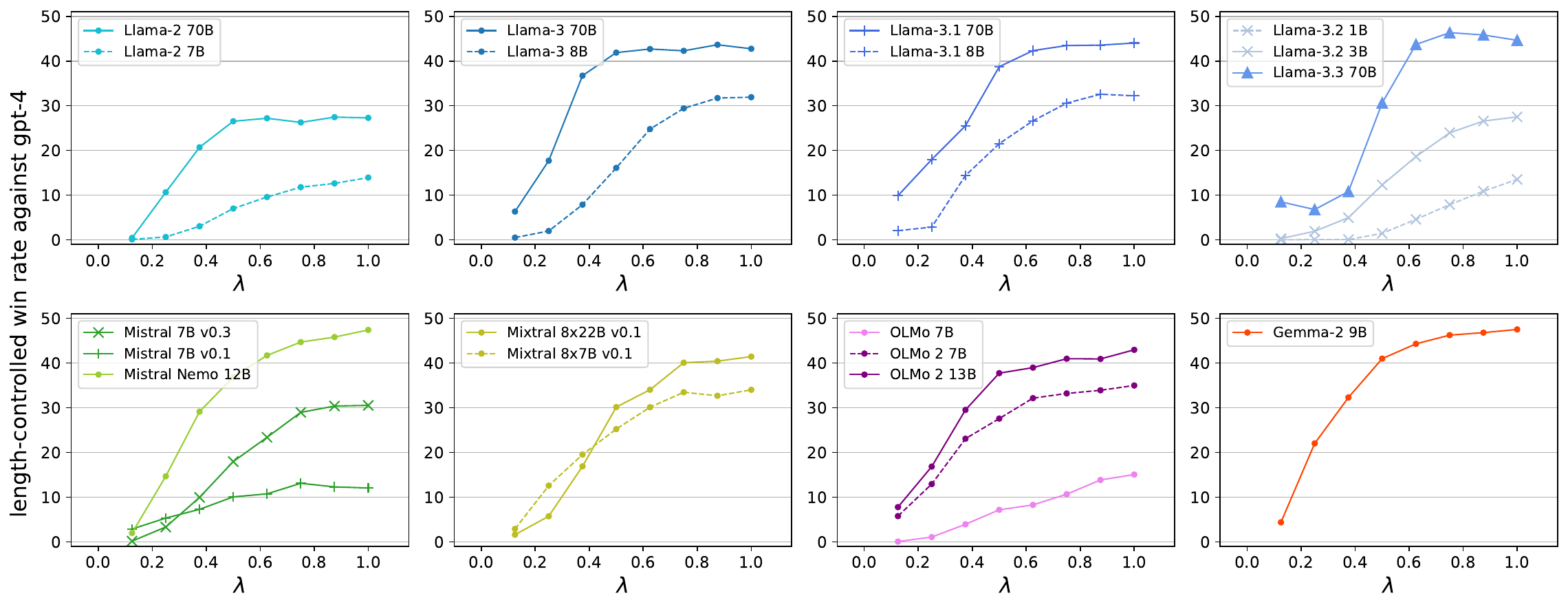}
    \caption{Performance on AlpacaEval 2.0: the length-controlled win rates of each partially adapted model $M_\lambda$ against GPT-4 Preview (11/06). }
    \label{fig:curves_alpaca_abs}
\end{figure*}

\begin{figure*}[h!]
    \centering
    \includegraphics[width=\textwidth]{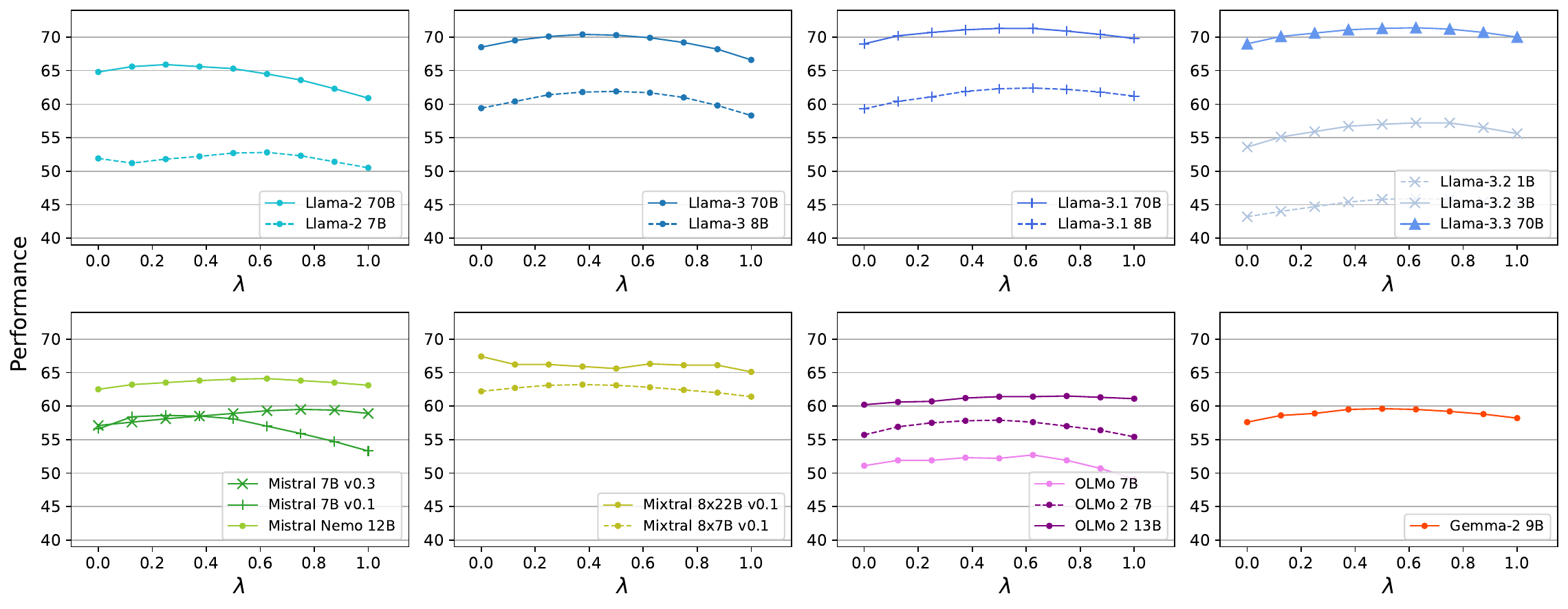}
    \caption{Performance on the in-context learning benchmark: absolute performance of each partially adapted model $M_\lambda$ for all the models we have tested. }
    \label{fig:curves_abs}
\end{figure*}

\newpage
\begin{sidewaystable}
\setlength{\tabcolsep}{2pt}
\footnotesize
\begin{center}

\begin{tabular}{ lll|llllllllllllllllllll } 
\toprule
Model & \rot{MMLU (base)} & \rotb{MMLU (inst.)} & \rot{MMLU} & \rot{ARC Challenge} & \rot{BBH (Algo)} & \rot{BBH (NLP)} & \rot{ConvFinQA} & \rot{DROP} & \rot{Headline} & \rot{Flue (NER)} & \rot{FiQA (SA)} & \rot{HellaSwag} & \rot{Natural Questions} & \rot{Pacific} & \rot{PIQA} & \rot{QuAC} & \rot{Natural Questions + Wiki} & \rot{Trivia QA + Wiki} & \rot{SQuAD} & \rot{TAT-QA} & \rot{Trivia QA} & \rot{Winogrande} \\
\midrule
Llama-3 70B & $79.1$ & $80.6$ & $81.0 ^{\color{gray} \sfrac{4}{8}}$ & $74.0 ^{\color{gray} \sfrac{7}{8}}$ & $57.9 ^{\color{gray} \sfrac{5}{8}}$ & $78.7 ^{\color{gray} \sfrac{2}{8}}$ & $69.9 ^{\color{gray} \sfrac{6}{8}}$ & $81.5 ^{\color{gray} \sfrac{3}{8}}$ & $91.6 ^{\color{gray} \sfrac{1}{8}}$ & $65.6 ^{\color{gray} 1}$ & $83.5 ^{\color{gray} \sfrac{6}{8}}$ & $85.6 ^{\color{gray} \sfrac{1}{8}}$ & $39.8 ^{\color{gray} \sfrac{3}{8}}$ & $67.0 ^{\color{gray} \sfrac{4}{8}}$ & $83.8 ^{\color{gray} \sfrac{2}{8}}$ & $60.3 ^{\color{gray} \sfrac{3}{8}}$ & $35.9 ^{\color{gray} \sfrac{5}{8}}$ & $80.6 ^{\color{gray} 0}$ & $53.2 ^{\color{gray} \sfrac{3}{8}}$ & $64.5 ^{\color{gray} \sfrac{2}{8}}$ & $82.0 ^{\color{gray} \sfrac{4}{8}}$ & $79.9 ^{\color{gray} \sfrac{2}{8}}$ \\
Llama-3 8B & $65.2$ & $66.7$ & $66.9 ^{\color{gray} \sfrac{7}{8}}$ & $64.8 ^{\color{gray} \sfrac{5}{8}}$ & $42.9 ^{\color{gray} \sfrac{7}{8}}$ & $66.8 ^{\color{gray} \sfrac{6}{8}}$ & $60.1 ^{\color{gray} \sfrac{5}{8}}$ & $64.0 ^{\color{gray} \sfrac{3}{8}}$ & $90.1 ^{\color{gray} \sfrac{4}{8}}$ & $67.0 ^{\color{gray} \sfrac{6}{8}}$ & $79.4 ^{\color{gray} \sfrac{7}{8}}$ & $79.4 ^{\color{gray} \sfrac{2}{8}}$ & $31.5 ^{\color{gray} \sfrac{2}{8}}$ & $55.0 ^{\color{gray} \sfrac{6}{8}}$ & $80.2 ^{\color{gray} \sfrac{4}{8}}$ & $52.0 ^{\color{gray} \sfrac{5}{8}}$ & $35.5 ^{\color{gray} \sfrac{3}{8}}$ & $74.3 ^{\color{gray} \sfrac{2}{8}}$ & $48.5 ^{\color{gray} \sfrac{7}{8}}$ & $52.4 ^{\color{gray} \sfrac{3}{8}}$ & $68.1 ^{\color{gray} \sfrac{5}{8}}$ & $73.0 ^{\color{gray} \sfrac{1}{8}}$ \\
Llama-3.1 70B & $78.8$ & $82.7$ & $82.7 ^{\color{gray} 1}$ & $72.1 ^{\color{gray} \sfrac{5}{8}}$ & $60.8 ^{\color{gray} \sfrac{7}{8}}$ & $80.5 ^{\color{gray} \sfrac{4}{8}}$ & $69.7 ^{\color{gray} \sfrac{3}{8}}$ & $82.3 ^{\color{gray} \sfrac{3}{8}}$ & $91.8 ^{\color{gray} \sfrac{2}{8}}$ & $65.6 ^{\color{gray} \sfrac{5}{8}}$ & $83.3 ^{\color{gray} \sfrac{5}{8}}$ & $85.8 ^{\color{gray} \sfrac{2}{8}}$ & $42.6 ^{\color{gray} \sfrac{4}{8}}$ & $67.1 ^{\color{gray} \sfrac{3}{8}}$ & $83.9 ^{\color{gray} \sfrac{3}{8}}$ & $59.4 ^{\color{gray} \sfrac{3}{8}}$ & $40.1 ^{\color{gray} \sfrac{7}{8}}$ & $81.9 ^{\color{gray} \sfrac{1}{8}}$ & $53.5 ^{\color{gray} \sfrac{4}{8}}$ & $67.1 ^{\color{gray} \sfrac{4}{8}}$ & $83.2 ^{\color{gray} \sfrac{1}{8}}$ & $81.0 ^{\color{gray} \sfrac{1}{8}}$ \\
Llama-3.1 8B & $65.6$ & $68.5$ & $68.5 ^{\color{gray} \sfrac{7}{8}}$ & $61.9 ^{\color{gray} \sfrac{6}{8}}$ & $46.9 ^{\color{gray} \sfrac{7}{8}}$ & $66.7 ^{\color{gray} \sfrac{5}{8}}$ & $59.9 ^{\color{gray} \sfrac{5}{8}}$ & $63.5 ^{\color{gray} \sfrac{4}{8}}$ & $89.6 ^{\color{gray} \sfrac{6}{8}}$ & $67.5 ^{\color{gray} \sfrac{5}{8}}$ & $78.6 ^{\color{gray} 1}$ & $79.5 ^{\color{gray} \sfrac{2}{8}}$ & $32.8 ^{\color{gray} \sfrac{4}{8}}$ & $55.3 ^{\color{gray} 1}$ & $80.1 ^{\color{gray} \sfrac{3}{8}}$ & $52.4 ^{\color{gray} \sfrac{5}{8}}$ & $39.8 ^{\color{gray} \sfrac{7}{8}}$ & $74.5 ^{\color{gray} \sfrac{1}{8}}$ & $44.5 ^{\color{gray} \sfrac{7}{8}}$ & $54.0 ^{\color{gray} \sfrac{6}{8}}$ & $68.1 ^{\color{gray} \sfrac{4}{8}}$ & $72.4 ^{\color{gray} \sfrac{2}{8}}$ \\
Llama-3.2 3B & $56.2$ & $61.2$ & $61.6 ^{\color{gray} \sfrac{6}{8}}$ & $55.6 ^{\color{gray} \sfrac{5}{8}}$ & $43.6 ^{\color{gray} \sfrac{6}{8}}$ & $62.3 ^{\color{gray} \sfrac{6}{8}}$ & $53.8 ^{\color{gray} \sfrac{4}{8}}$ & $53.0 ^{\color{gray} \sfrac{5}{8}}$ & $89.7 ^{\color{gray} \sfrac{5}{8}}$ & $60.7 ^{\color{gray} \sfrac{4}{8}}$ & $78.3 ^{\color{gray} \sfrac{7}{8}}$ & $74.0 ^{\color{gray} \sfrac{2}{8}}$ & $28.9 ^{\color{gray} \sfrac{3}{8}}$ & $48.3 ^{\color{gray} \sfrac{7}{8}}$ & $77.8 ^{\color{gray} \sfrac{3}{8}}$ & $48.9 ^{\color{gray} 1}$ & $37.9 ^{\color{gray} \sfrac{3}{8}}$ & $68.9 ^{\color{gray} \sfrac{3}{8}}$ & $44.8 ^{\color{gray} 1}$ & $49.6 ^{\color{gray} \sfrac{5}{8}}$ & $57.1 ^{\color{gray} \sfrac{3}{8}}$ & $67.2 ^{\color{gray} \sfrac{1}{8}}$ \\
Llama-3.2 1B & $37.7$ & $45.2$ & $45.2 ^{\color{gray} \sfrac{7}{8}}$ & $44.5 ^{\color{gray} \sfrac{5}{8}}$ & $29.2 ^{\color{gray} \sfrac{6}{8}}$ & $53.3 ^{\color{gray} \sfrac{3}{8}}$ & $38.1 ^{\color{gray} \sfrac{5}{8}}$ & $30.9 ^{\color{gray} \sfrac{3}{8}}$ & $83.5 ^{\color{gray} \sfrac{5}{8}}$ & $58.3 ^{\color{gray} 1}$ & $72.3 ^{\color{gray} 1}$ & $63.0 ^{\color{gray} \sfrac{2}{8}}$ & $18.8 ^{\color{gray} \sfrac{4}{8}}$ & $30.4 ^{\color{gray} \sfrac{7}{8}}$ & $76.0 ^{\color{gray} \sfrac{2}{8}}$ & $40.0 ^{\color{gray} \sfrac{5}{8}}$ & $31.6 ^{\color{gray} \sfrac{1}{8}}$ & $54.4 ^{\color{gray} \sfrac{3}{8}}$ & $33.5 ^{\color{gray} 1}$ & $35.1 ^{\color{gray} 1}$ & $37.2 ^{\color{gray} \sfrac{4}{8}}$ & $60.0 ^{\color{gray} 0}$ \\
Llama-3.3 70B & $78.9$ & $82.7$ & $82.7 ^{\color{gray} 1}$ & $73.1 ^{\color{gray} \sfrac{5}{8}}$ & $61.1 ^{\color{gray} 1}$ & $80.5 ^{\color{gray} \sfrac{4}{8}}$ & $69.5 ^{\color{gray} \sfrac{4}{8}}$ & $82.2 ^{\color{gray} \sfrac{4}{8}}$ & $92.0 ^{\color{gray} \sfrac{4}{8}}$ & $66.7 ^{\color{gray} \sfrac{6}{8}}$ & $83.2 ^{\color{gray} \sfrac{7}{8}}$ & $85.7 ^{\color{gray} \sfrac{2}{8}}$ & $40.6 ^{\color{gray} \sfrac{4}{8}}$ & $67.4 ^{\color{gray} \sfrac{5}{8}}$ & $84.2 ^{\color{gray} \sfrac{4}{8}}$ & $58.9 ^{\color{gray} \sfrac{5}{8}}$ & $38.1 ^{\color{gray} \sfrac{7}{8}}$ & $81.7 ^{\color{gray} \sfrac{2}{8}}$ & $59.4 ^{\color{gray} \sfrac{5}{8}}$ & $66.4 ^{\color{gray} \sfrac{3}{8}}$ & $83.4 ^{\color{gray} \sfrac{2}{8}}$ & $80.8 ^{\color{gray} \sfrac{1}{8}}$ \\
Llama-2 70B & $69.2$ & $63.6$ & $70.2 ^{\color{gray} \sfrac{2}{8}}$ & $67.7 ^{\color{gray} \sfrac{2}{8}}$ & $53.3 ^{\color{gray} \sfrac{2}{8}}$ & $71.7 ^{\color{gray} \sfrac{2}{8}}$ & $61.7 ^{\color{gray} \sfrac{4}{8}}$ & $71.6 ^{\color{gray} \sfrac{2}{8}}$ & $91.2 ^{\color{gray} \sfrac{2}{8}}$ & $62.7 ^{\color{gray} 1}$ & $81.8 ^{\color{gray} \sfrac{6}{8}}$ & $83.4 ^{\color{gray} \sfrac{2}{8}}$ & $42.8 ^{\color{gray} \sfrac{1}{8}}$ & $56.3 ^{\color{gray} \sfrac{2}{8}}$ & $82.8 ^{\color{gray} \sfrac{3}{8}}$ & $54.7 ^{\color{gray} \sfrac{4}{8}}$ & $39.5 ^{\color{gray} 0}$ & $79.5 ^{\color{gray} \sfrac{1}{8}}$ & $50.6 ^{\color{gray} \sfrac{6}{8}}$ & $55.9 ^{\color{gray} \sfrac{4}{8}}$ & $80.3 ^{\color{gray} 0}$ & $79.6 ^{\color{gray} \sfrac{1}{8}}$ \\
Llama-2 7B & $44.1$ & $47.3$ & $47.9 ^{\color{gray} \sfrac{6}{8}}$ & $55.2 ^{\color{gray} \sfrac{3}{8}}$ & $33.6 ^{\color{gray} 0}$ & $57.0 ^{\color{gray} \sfrac{6}{8}}$ & $43.2 ^{\color{gray} \sfrac{5}{8}}$ & $42.2 ^{\color{gray} \sfrac{5}{8}}$ & $86.9 ^{\color{gray} \sfrac{7}{8}}$ & $64.6 ^{\color{gray} \sfrac{2}{8}}$ & $74.1 ^{\color{gray} \sfrac{6}{8}}$ & $75.9 ^{\color{gray} \sfrac{3}{8}}$ & $28.5 ^{\color{gray} 0}$ & $38.7 ^{\color{gray} \sfrac{7}{8}}$ & $78.8 ^{\color{gray} \sfrac{4}{8}}$ & $43.8 ^{\color{gray} \sfrac{5}{8}}$ & $33.6 ^{\color{gray} 0}$ & $67.4 ^{\color{gray} 0}$ & $38.8 ^{\color{gray} 0}$ & $39.6 ^{\color{gray} \sfrac{5}{8}}$ & $59.7 ^{\color{gray} \sfrac{2}{8}}$ & $67.9 ^{\color{gray} \sfrac{4}{8}}$ \\
Gemma-2 9B & $72.0$ & $72.6$ & $73.6 ^{\color{gray} \sfrac{3}{8}}$ & $71.6 ^{\color{gray} \sfrac{7}{8}}$ & $48.5 ^{\color{gray} \sfrac{3}{8}}$ & $72.8 ^{\color{gray} \sfrac{5}{8}}$ & $52.6 ^{\color{gray} \sfrac{5}{8}}$ & $54.7 ^{\color{gray} \sfrac{4}{8}}$ & $90.4 ^{\color{gray} \sfrac{2}{8}}$ & $45.7 ^{\color{gray} \sfrac{6}{8}}$ & $82.4 ^{\color{gray} 1}$ & $80.7 ^{\color{gray} \sfrac{4}{8}}$ & $34.7 ^{\color{gray} \sfrac{3}{8}}$ & $39.9 ^{\color{gray} \sfrac{3}{8}}$ & $82.1 ^{\color{gray} \sfrac{4}{8}}$ & $33.3 ^{\color{gray} \sfrac{6}{8}}$ & $42.7 ^{\color{gray} \sfrac{1}{8}}$ & $76.1 ^{\color{gray} \sfrac{3}{8}}$ & $41.2 ^{\color{gray} \sfrac{7}{8}}$ & $41.0 ^{\color{gray} 1}$ & $67.4 ^{\color{gray} \sfrac{2}{8}}$ & $75.5 ^{\color{gray} \sfrac{1}{8}}$ \\
Mixtral 8x22B v0.1 & $76.2$ & $76.6$ & $76.7 ^{\color{gray} \sfrac{6}{8}}$ & $71.5 ^{\color{gray} 1}$ & $55.6 ^{\color{gray} 0}$ & $74.7 ^{\color{gray} 0}$ & $72.6 ^{\color{gray} \sfrac{6}{8}}$ & $78.4 ^{\color{gray} \sfrac{4}{8}}$ & $89.8 ^{\color{gray} 0}$ & $66.7 ^{\color{gray} 0}$ & $82.9 ^{\color{gray} 1}$ & $82.9 ^{\color{gray} \sfrac{7}{8}}$ & $42.1 ^{\color{gray} \sfrac{2}{8}}$ & $64.3 ^{\color{gray} 0}$ & $75.5 ^{\color{gray} \sfrac{5}{8}}$ & $56.1 ^{\color{gray} \sfrac{6}{8}}$ & $42.1 ^{\color{gray} \sfrac{2}{8}}$ & $82.0 ^{\color{gray} \sfrac{2}{8}}$ & $49.5 ^{\color{gray} \sfrac{7}{8}}$ & $64.0 ^{\color{gray} \sfrac{5}{8}}$ & $82.4 ^{\color{gray} 0}$ & $68.7 ^{\color{gray} \sfrac{7}{8}}$ \\
Mixtral 8x7B v0.1 & $69.1$ & $69.3$ & $70.0 ^{\color{gray} \sfrac{3}{8}}$ & $70.8 ^{\color{gray} 1}$ & $48.6 ^{\color{gray} \sfrac{3}{8}}$ & $71.5 ^{\color{gray} \sfrac{5}{8}}$ & $64.0 ^{\color{gray} \sfrac{7}{8}}$ & $67.9 ^{\color{gray} \sfrac{3}{8}}$ & $89.2 ^{\color{gray} \sfrac{6}{8}}$ & $64.3 ^{\color{gray} \sfrac{7}{8}}$ & $79.2 ^{\color{gray} \sfrac{7}{8}}$ & $81.8 ^{\color{gray} 1}$ & $37.9 ^{\color{gray} 0}$ & $56.2 ^{\color{gray} \sfrac{2}{8}}$ & $76.6 ^{\color{gray} \sfrac{7}{8}}$ & $50.6 ^{\color{gray} \sfrac{6}{8}}$ & $40.5 ^{\color{gray} \sfrac{2}{8}}$ & $78.5 ^{\color{gray} 0}$ & $44.3 ^{\color{gray} \sfrac{3}{8}}$ & $56.7 ^{\color{gray} \sfrac{3}{8}}$ & $77.5 ^{\color{gray} \sfrac{4}{8}}$ & $65.1 ^{\color{gray} 0}$ \\
Mistral 7B v0.1 & $56.9$ & $49.9$ & $57.5 ^{\color{gray} \sfrac{2}{8}}$ & $64.1 ^{\color{gray} \sfrac{2}{8}}$ & $45.5 ^{\color{gray} \sfrac{2}{8}}$ & $65.0 ^{\color{gray} \sfrac{2}{8}}$ & $54.5 ^{\color{gray} \sfrac{2}{8}}$ & $58.8 ^{\color{gray} \sfrac{4}{8}}$ & $90.5 ^{\color{gray} \sfrac{3}{8}}$ & $61.5 ^{\color{gray} 0}$ & $78.2 ^{\color{gray} \sfrac{6}{8}}$ & $77.0 ^{\color{gray} \sfrac{1}{8}}$ & $33.1 ^{\color{gray} \sfrac{1}{8}}$ & $46.6 ^{\color{gray} \sfrac{1}{8}}$ & $74.4 ^{\color{gray} \sfrac{2}{8}}$ & $51.6 ^{\color{gray} \sfrac{7}{8}}$ & $37.1 ^{\color{gray} \sfrac{5}{8}}$ & $69.8 ^{\color{gray} \sfrac{2}{8}}$ & $47.4 ^{\color{gray} \sfrac{7}{8}}$ & $50.4 ^{\color{gray} \sfrac{2}{8}}$ & $65.8 ^{\color{gray} \sfrac{1}{8}}$ & $62.9 ^{\color{gray} \sfrac{2}{8}}$ \\
Mistral 7B v0.3 & $59.7$ & $59.9$ & $60.0 ^{\color{gray} \sfrac{2}{8}}$ & $66.0 ^{\color{gray} \sfrac{7}{8}}$ & $44.3 ^{\color{gray} \sfrac{3}{8}}$ & $65.5 ^{\color{gray} \sfrac{5}{8}}$ & $56.0 ^{\color{gray} \sfrac{6}{8}}$ & $55.6 ^{\color{gray} \sfrac{5}{8}}$ & $90.4 ^{\color{gray} 1}$ & $64.7 ^{\color{gray} \sfrac{5}{8}}$ & $80.3 ^{\color{gray} 1}$ & $79.5 ^{\color{gray} 1}$ & $28.7 ^{\color{gray} \sfrac{4}{8}}$ & $48.6 ^{\color{gray} \sfrac{4}{8}}$ & $75.7 ^{\color{gray} 1}$ & $54.9 ^{\color{gray} \sfrac{7}{8}}$ & $36.1 ^{\color{gray} \sfrac{4}{8}}$ & $71.8 ^{\color{gray} \sfrac{6}{8}}$ & $44.3 ^{\color{gray} 1}$ & $49.4 ^{\color{gray} \sfrac{6}{8}}$ & $66.0 ^{\color{gray} \sfrac{6}{8}}$ & $64.2 ^{\color{gray} \sfrac{6}{8}}$ \\
Mistral Nemo 2407 & $68.4$ & $69.1$ & $69.4 ^{\color{gray} \sfrac{4}{8}}$ & $63.2 ^{\color{gray} \sfrac{5}{8}}$ & $46.5 ^{\color{gray} \sfrac{5}{8}}$ & $70.6 ^{\color{gray} \sfrac{5}{8}}$ & $64.1 ^{\color{gray} \sfrac{6}{8}}$ & $72.1 ^{\color{gray} \sfrac{4}{8}}$ & $88.1 ^{\color{gray} 0}$ & $64.4 ^{\color{gray} 1}$ & $80.4 ^{\color{gray} 1}$ & $81.6 ^{\color{gray} \sfrac{4}{8}}$ & $32.2 ^{\color{gray} \sfrac{3}{8}}$ & $56.4 ^{\color{gray} 1}$ & $81.8 ^{\color{gray} \sfrac{5}{8}}$ & $57.9 ^{\color{gray} 1}$ & $38.6 ^{\color{gray} \sfrac{3}{8}}$ & $74.6 ^{\color{gray} \sfrac{1}{8}}$ & $49.1 ^{\color{gray} \sfrac{5}{8}}$ & $58.1 ^{\color{gray} \sfrac{5}{8}}$ & $72.1 ^{\color{gray} 0}$ & $75.7 ^{\color{gray} \sfrac{4}{8}}$ \\
OLMo 7B 0724 & $52.0$ & $53.0$ & $54.4 ^{\color{gray} \sfrac{4}{8}}$ & $51.4 ^{\color{gray} \sfrac{6}{8}}$ & $35.9 ^{\color{gray} \sfrac{3}{8}}$ & $57.4 ^{\color{gray} 0}$ & $35.8 ^{\color{gray} \sfrac{5}{8}}$ & $53.1 ^{\color{gray} \sfrac{4}{8}}$ & $88.7 ^{\color{gray} \sfrac{2}{8}}$ & $63.4 ^{\color{gray} 0}$ & $74.4 ^{\color{gray} \sfrac{5}{8}}$ & $79.5 ^{\color{gray} 1}$ & $33.0 ^{\color{gray} 0}$ & $36.8 ^{\color{gray} \sfrac{5}{8}}$ & $79.9 ^{\color{gray} \sfrac{3}{8}}$ & $32.0 ^{\color{gray} \sfrac{5}{8}}$ & $35.2 ^{\color{gray} \sfrac{3}{8}}$ & $64.3 ^{\color{gray} 0}$ & $41.0 ^{\color{gray} \sfrac{1}{8}}$ & $40.9 ^{\color{gray} \sfrac{5}{8}}$ & $53.7 ^{\color{gray} 0}$ & $68.6 ^{\color{gray} \sfrac{2}{8}}$ \\
OLMo 2 13B 1124 & $67.0$ & $66.4$ & $67.0 ^{\color{gray} 0}$ & $67.3 ^{\color{gray} 1}$ & $39.0 ^{\color{gray} 1}$ & $66.5 ^{\color{gray} 1}$ & $50.3 ^{\color{gray} \sfrac{6}{8}}$ & $73.4 ^{\color{gray} 0}$ & $87.0 ^{\color{gray} 0}$ & $63.5 ^{\color{gray} \sfrac{7}{8}}$ & $77.4 ^{\color{gray} 1}$ & $85.9 ^{\color{gray} 1}$ & $39.8 ^{\color{gray} 0}$ & $51.6 ^{\color{gray} \sfrac{5}{8}}$ & $81.9 ^{\color{gray} 1}$ & $48.5 ^{\color{gray} \sfrac{6}{8}}$ & $42.8 ^{\color{gray} 0}$ & $71.7 ^{\color{gray} 0}$ & $43.7 ^{\color{gray} 1}$ & $51.3 ^{\color{gray} \sfrac{6}{8}}$ & $68.2 ^{\color{gray} 0}$ & $75.7 ^{\color{gray} \sfrac{1}{8}}$ \\
OLMo 2 7B 1124 & $62.7$ & $61.2$ & $63.1 ^{\color{gray} \sfrac{3}{8}}$ & $64.2 ^{\color{gray} \sfrac{6}{8}}$ & $33.9 ^{\color{gray} \sfrac{6}{8}}$ & $63.0 ^{\color{gray} \sfrac{7}{8}}$ & $44.3 ^{\color{gray} \sfrac{6}{8}}$ & $63.2 ^{\color{gray} \sfrac{1}{8}}$ & $85.8 ^{\color{gray} \sfrac{4}{8}}$ & $59.5 ^{\color{gray} 0}$ & $74.7 ^{\color{gray} 0}$ & $83.6 ^{\color{gray} 1}$ & $32.5 ^{\color{gray} \sfrac{3}{8}}$ & $44.3 ^{\color{gray} \sfrac{4}{8}}$ & $80.7 ^{\color{gray} \sfrac{7}{8}}$ & $46.9 ^{\color{gray} \sfrac{4}{8}}$ & $41.2 ^{\color{gray} \sfrac{1}{8}}$ & $67.4 ^{\color{gray} \sfrac{2}{8}}$ & $43.6 ^{\color{gray} \sfrac{3}{8}}$ & $46.2 ^{\color{gray} \sfrac{2}{8}}$ & $64.5 ^{\color{gray} \sfrac{1}{8}}$ & $71.3 ^{\color{gray} \sfrac{3}{8}}$ \\
\bottomrule
\end{tabular}

\end{center}
\caption{
Fine-grained results for each model and each dataset in our benchmark. Each entry reports the best score achieved and the value of $\lambda$ at which such score was achieved in the format ${\textrm{score}^\lambda}$. We also report MMLU performances for the base and instruct version of every model in the first two columns.}
\label{tab:results_nonagg}

\end{sidewaystable}

\end{document}